\documentclass[conference]{IEEEtran}
\IEEEoverridecommandlockouts
\usepackage{algorithm}      
\usepackage{algpseudocode}  

\usepackage{graphicx}
\usepackage{subcaption}
\usepackage{amsmath} 
\usepackage{amssymb} 
\usepackage{booktabs}
\usepackage{cite}
\usepackage{amsmath,amssymb,amsfonts}
\usepackage{graphicx}
\usepackage{textcomp}
\usepackage{xcolor}
\usepackage{amsmath}
\usepackage{hyperref}
\def\BibTeX{{\rm B\kern-.05em{\sc i\kern-.025em b}\kern-.08em
    T\kern-.1667em\lower.7ex\hbox{E}\kern-.125emX}}
\begin{document}

\title{3D Graph Attention Networks for High-Fidelity Pediatric Glioma Segmentation\\
}

\author{\IEEEauthorblockN{Harish Thangaraj\textsuperscript{1}, Diya Katariya\textsuperscript{2}, 
Eshaan Joshi\textsuperscript{3}, Sangeetha N.\textsuperscript{4}}
\IEEEauthorblockA{\textit{School of Computer Science and Engineering} \\
\textit{Vellore Institute of Technology, Chennai}\\
Chennai, India \\
\textsuperscript{1}harish.thangaraj03@outlook.com, 
\textsuperscript{2}diyakatariya0907@gmail.com, 
\textsuperscript{3}eshaanjoshi.work@gmail.com,
\textsuperscript{4}n.sangeetha@vit.ac.in}
}

\maketitle

\begin{abstract}
Pediatric brain tumors, particularly gliomas, represent a significant cause of cancer-related mortality in children, with complex, infiltrative growth patterns that complicate treatment. Early, accurate segmentation of these tumors in neuroimaging data is crucial for effective diagnosis and intervention planning. This study presents a novel 3D U-Net architecture with a spatial attention mechanism tailored for automated segmentation of pediatric gliomas. Using the BraTS pediatric glioma dataset with multiparametric MRI data, the proposed model captures multi-scale features and selectively attends to tumor-relevant regions, enhancing segmentation precision and reducing interference from surrounding tissue. The model’s performance is quantitatively evaluated using the Dice similarity coefficient and HD95, demonstrating improved delineation of complex glioma structures. This approach offers a promising advancement in automating pediatric glioma segmentation, with the potential to improve clinical decision-making and outcomes.
\end{abstract}


\section{Introduction}

Pediatric brain tumors are the leading cause of cancer-related mortality in children. Among these, gliomas—tumors arising from glial cells that normally support neurons in the central nervous system—pose a significant clinical challenge. Gliomas frequently infiltrate sensitive regions of the brain, complicating surgical resection, and can spread into adjacent tissues, thus increasing their malignancy and life-threatening potential. These challenges underscore the urgent need for timely and accurate tumor detection, diagnosis, and characterization.

Neuroimaging techniques such as positron emission tomography (PET), magnetic resonance imaging (MRI), and computed tomography (CT) have been widely employed to identify subtle anatomical anomalies in the brain. In current clinical practice, radiologists typically review each two-dimensional (2D) image slice of a brain scan, manually delineating boundaries around suspected tumor regions. The resulting 2D contours can then be reconstructed into three-dimensional (3D) volumes to assess tumor morphology, volume, and subregional distinctions. Integrating this process with deep learning architectures has the potential to automate and expedite tumor segmentation, thereby reducing human error and increasing both efficiency and accuracy.

The Brain Tumor Segmentation (BraTS) challenge, hosted by the Annual Medical Image Computing and Computer Assisted Interventions (MICCAI) conference, provides a platform for researchers to develop and benchmark state-of-the-art segmentation algorithms using clinically acquired, multi-parametric MRI datasets. Building upon these advances, this paper introduces a novel U-Net architecture enhanced with a spatial attention mechanism to accurately segment and analyze tumors from the BraTS pediatric glioma dataset. This dataset includes images of diffuse midline glioma (DMG) and diffuse intrinsic pontine glioma (DIPG), accompanied by ground truth annotations for quantitative evaluation of predicted segmentations.

The proposed U-Net model employs an encoder-decoder framework to capture multi-scale features and preserve spatial relationships within 3D MRI scans. By incorporating a spatial attention module, the model selectively emphasizes tumor-relevant regions in the MRI while mitigating the influence of irrelevant background structures. To quantitatively assess the accuracy of the proposed segmentation approach, we employ the Dice Similarity Coefficient (DSC) and the 95th percentile Hausdorff distance (HD95).

\renewcommand{\arraystretch}{1.3} 

\begin{table}[h!]
\centering
\caption{Training Parameters for Model Training}

\begin{tabular}{@{}ll@{}}
\toprule
\textbf{Parameter} & \textbf{Value} \\ 
\midrule
is\_debugging & False \\
all\_samples\_as\_train & False \\
fold & 4 \\
seed & 42 \\
max\_epochs & 75 \\
mednext\_size & B \\
mednext\_ksize & 3 \\
mednext\_ckpt & None \\
deep\_sup & True \\
batch\_size & 1 \\
sw\_batch\_size & 2 \\
num\_workers & 4 \\
roi\_x & 128 \\
roi\_y & 128 \\
roi\_z & 128 \\
infer\_overlap & 0.5 \\
aug\_type & 1 \\
loss\_type & 3 \\
mean\_batch & True \\
learning rate (lr) & $3 \times 10^{-4}$ \\
weight decay & $1 \times 10^{-6}$ \\
lr\_scheduler & cosine-with-warmup \\
n\_gpus & 1 \\
pin\_memory & True \\
check\_val\_every\_n\_epoch & 1 \\
precision & 16 \\
amp\_backend & native \\
accumulate\_grad\_batches & 4 \\
\bottomrule
\end{tabular}
\label{table:training_parameters}
\end{table}

\section{Related Works}

\subsection{Advancements in Brain Tumor Segmentation Techniques}

Various advancements have been made in brain tumor segmentation, focusing on enhancing diagnostic precision and automation. Maani et al \cite{Maani2024} demonstrated the efficacy of combining SegResNet and MedNeXt in improving segmentation accuracy for adult and pediatric gliomas, achieving a Dice score of 0.8313 in the BraTS 2023 challenge. Funakoshi et al \cite{Funakoshi2021} emphasized the role of next-generation sequencing (NGS) and liquid biopsy in identifying genetic mutations like BRAF V600E, paving the way for personalized treatments. Dequidt et al. (2021) proposed a support vector machine (SVM)-based classifier for glioma grading, achieving 82.4\% accuracy, while Jiang et al \cite{Jiang2020} implemented a two-stage cascaded U-Net for precise tumor segmentation, earning first place in the BraTS 2019 challenge. Goswami et al. (2013) \cite{Goswami2013} introduced an unsupervised neural network using ICA and SOM clustering for non-invasive tumor classification. Pedada et al. (2023) \cite{Pedada2023} improved segmentation using a residual-based U-Net, achieving 93.40\% accuracy on BraTS datasets. Surveys by Liu et al. (2014) \cite{Liu2014} and Biratu et al. (2021) \cite{Biratu2021} highlighted the limitations of traditional methods, advocating for machine learning approaches for improved segmentation. Zhang et al. (2024) \cite{Zhang2024} proposed a multi-attention gate in U-Net to enhance low-grade glioma segmentation accuracy to 99.7\%. Patel et al. (2023) \cite{Patel2023} employed graph attention networks (GATs) for multiclass segmentation, outperforming baseline methods with mean Dice scores above 0.79. Khan et al. (2023) \cite{Khan2023}introduced a hybrid residual U-Net incorporating attention modules, achieving superior performance on BraTS datasets. Bakas et al. (2018) \cite{Bakas2018} identified consensus-based segmentation as a promising approach for robust tumor delineation. Kofler et al. (2020)\cite{Kofler2020} showcased the BraTS Toolkit's ability to standardize and improve segmentation accuracy through consensus methods. Lastly, Kazeroon et al. (2024) \cite{Kazerooni2024} and Nikam et al. (2022) \cite{Nikam2022}highlighted challenges in pediatric brain tumor segmentation and the potential of radiogenomics for non-invasive diagnostic advancements. Additionally, Hashmi et al. \cite{hashmi2024optimizingbraintumorsegmentation} introduced an advanced MedNeXt-based architecture for the BraTS 2024 SSA and Pediatric Tumor tasks, demonstrating exceptional segmentation performance with Dice Similarity Coefficients (DSC) of 0.896 and 0.830 on the SSA and Pediatric datasets, respectively. Their methodology effectively addressed distribution shifts in data, such as lower-quality MRIs and varying patient demographics, showcasing the robustness of the MedNeXt framework and the schedule-free optimizer.

\section{Dataset and Preprocessing}
The Brain Tumor Segmentation (BraTS) dataset is a widely used collection of annotated MRI scans designed to benchmark and advance methods for brain tumor segmentation, particularly gliomas. The dataset is curated by experts in neuro-oncology and medical imaging, making it a reliable resource for research and development in this field. The dataset focuses on the segmentation of gliomas, which are among the most aggressive brain tumors. It aims to provide researchers with standardized data to develop models for automated segmentation and analysis.

\subsection{BraTs Dataset}
The dataset utilized in this study is the BraTS Pediatric Glioma Dataset, sourced from the Brain Tumor Segmentation (BraTS) Challenge. It comprises multiparametric MRI scans of pediatric brain tumors, including diffuse midline gliomas (DMG) and diffuse intrinsic pontine gliomas (DIPG). The dataset features four MRI modalities: T1-weighted, T1 contrast-enhanced (T1ce), T2-weighted, and FLAIR, each capturing distinct tumor characteristics. T1-weighted scans provide structural details, T1ce highlights active tumor growth, T2 reflects water content and swelling, and FLAIR identifies edema. Ground truth annotations, meticulously labeled by expert neuroradiologists, classify tumor regions into enhancing tumors (ET), peritumoral edema (ED), and necrotic or non-enhancing tumor core (NCR/NET). The dataset offers standardized 3D volumes and includes both high-grade gliomas (HGG) and low-grade gliomas (LGG), with 220 HGG and 54 LGG cases in the BraTS 2016 training set. A blind testing set of 191 cases ensures unbiased model evaluation. Stored in NIfTI format, the dataset includes segmentation labels as binary masks for enhancing tumors, edema, and necrosis, supporting accurate and consistent tumor analysis.
\begin{figure}[h]
    \centering
    \includegraphics[width=0.5\textwidth]{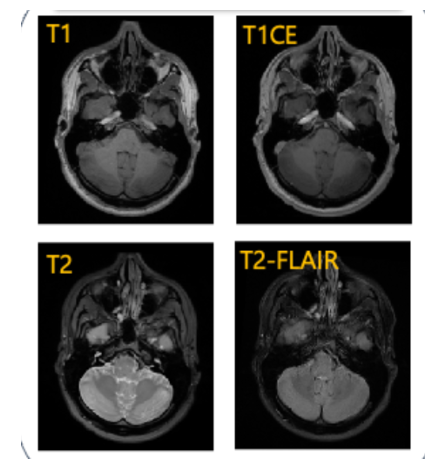}
    \caption{Overview of the BraTs Dataset}
    \label{fig:mednext_arch}
\end{figure}
\subsection{Dataset Preparation and Preprocessing}

\subsubsection{Standardization and Cropping}
The raw MRI scans were standardized to ensure uniform and clean input for the model. Intensity normalization was applied across modalities to maintain consistent brightness levels. Each scan was resized to 128 × 128 × 128 voxels, preserving tumor structural details while cropping out unnecessary background regions to focus on tumor-prone areas.

\subsubsection{Data Augmentation}
Augmentation techniques, such as random rotations, flips, and brightness adjustments, were implemented to enhance the model’s robustness. These variations simulated real-world conditions, increasing dataset size and improving the model’s ability to generalize, particularly in the context of pediatric tumor variability.

\subsubsection{Dataset Splitting}
The dataset was divided into training, validation, and testing subsets with balanced representation of tumor types, including high-grade gliomas (HGG) and low-grade gliomas (LGG). This ensured effective learning and unbiased evaluation during testing.

\subsubsection{Binary Mask Creation}
Segmentation labels were converted into binary masks to simplify the model's task. These masks distinguished tumor regions from surrounding brain tissue, with an additional binary channel added to enhance contrast between tumor and background regions.

\subsubsection{Validation of Preprocessing}
All preprocessing steps were validated to ensure dataset integrity. Alignment between MRI scans and ground truth annotations was cross-checked, preserving structural and spatial integrity of tumor regions during resizing and augmentation.

\subsubsection{Conclusion}
These systematic preprocessing and augmentation techniques ensured high-quality inputs for the model, enhancing its accuracy and reliability in segmenting pediatric gliomas and aligning the dataset with the study’s objectives.

\section{Proposed Methodology}

\subsection{U-Net Model}
The U-Net model is well-regarded for medical image segmentation due to its ability to preserve spatial information through skip connections. It features an encoder-decoder structure where the encoding path captures key features by progressively downsampling the image, while the decoding path reconstructs it to the original size, producing a segmented output. Skip connections between corresponding layers in the encoder and decoder help retain spatial details, making the model effective for accurately segmenting complex structures like tumors. By incorporating graph-attention layers, the proposed model further enhances focus on critical tumor regions, refining segmentation accuracy by distinguishing tumors from surrounding tissue. This combination captures fine-grained details, improving the quality and interpretability of segmented images for pediatric glioma detection.

\subsection{MedNeXt Architecture}
MedNeXt is a ConvNeXt-based U-Net architecture designed specifically for 3D medical imaging tasks. It employs convolutional layers for robust feature extraction and uses ReLU activation functions to introduce non-linearity. The architecture leverages downsampling and upsampling operations to effectively capture multi-scale features. To support gradient flow during training, skip connections are integrated to retain important spatial details across the network. MedNeXt also incorporates a cross-attention mechanism to selectively weigh and merge features from different sources, enhancing the focus on relevant anatomical structures. While precise, the reliance on channel attention limits its spatial selectivity, which motivates the use of the spatial attention mechanism in the proposed model.

\begin{figure}[h]
    \centering
    \includegraphics[width=0.5\textwidth]{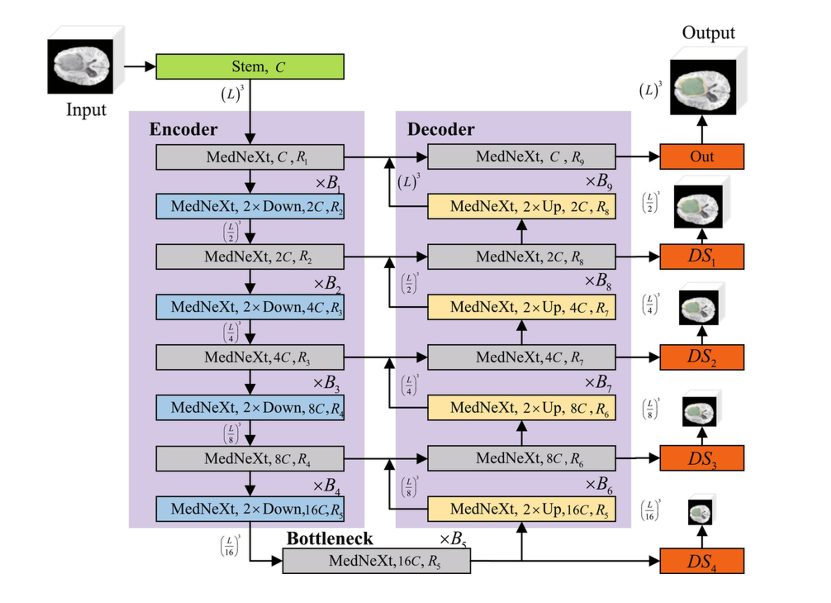}
    \caption{Overview of the MedNext architecture for 3D medical imaging.}
    \label{fig:mednext_arch}
\end{figure}

\begin{algorithm}[h]
\caption{MedNeXt Architecture for 3D Medical Imaging}
\label{alg:mednext}
\begin{algorithmic}[1]
\Require 3D medical image $I$
\State Initialize model parameters
\State Initialize optimizer
\State $F \gets I$ \Comment{Initial feature}
\For{\textbf{each} convolutional block \textbf{do}}
    \State $F \gets \text{Conv}(F)$
    \State $F \gets \text{ReLU}(F)$
    \State $F \gets \text{Downsample}(F)$
\EndFor
\For{\textbf{each} upsampling block \textbf{do}}
    \State $F \gets \text{Upsample}(F)$
    \State $F \gets \text{Concat}(F, \text{Skip})$
    \State $F \gets \text{Conv}(F)$
\EndFor
\State $F \gets \text{CrossAttention}(F, \text{Context})$
\State Output $\gets \text{SegmentationMap}(F)$
\State \Return Output
\end{algorithmic}
\end{algorithm}

\subsection{Graph-Based Spatial Attention Mechanism}
The proposed model integrates a graph-based spatial attention mechanism to dynamically enhance the focus on diagnostically relevant regions in 3D medical images. This mechanism uses a Graph Cross Attention (GCA) module that operates on a 3D graph representation of the image. 

The mechanism begins by constructing a graph structure where each voxel in the image acts as a node, and edges are established between neighboring voxels. Formally, let $x \in \mathbb{R}^{B \times C \times D \times H \times W}$ represent the input tensor, where $B$ is the batch size, $C$ the number of channels, and $D$, $H$, and $W$ denote depth, height, and width, respectively. For each voxel, a neighborhood connection is defined using:
\[
\text{edge\_index} = \{(i, j) \mid j \in \text{Neighbors}(i)\},
\]
where $i$ and $j$ are indices of connected nodes.

The module applies separate graph convolutional layers for query, key, and value projections:

\begin{equation}
\begin{split}
Q & = \text{GATv2Conv}(x), \\
K & = \text{GATv2Conv}(x), \\
V & = \text{GATv2Conv}(x).
\end{split}
\end{equation}

where $\text{GATv2Conv}$ represents the Graph Attention Network convolution. These projections reduce or preserve the channel dimensions as needed.

The attention weights are computed using a scaled dot product:
\[
\text{Energy}_{ijk} = Q_{ij} \cdot K_{ik},
\]
where $Q_{ij}$ and $K_{ik}$ represent the query and key features for nodes $i$ and $k$. These are normalized using the softmax function:
\[
\text{Attention}_{ijk} = \frac{\exp(\text{Energy}_{ijk})}{\sum_{k'} \exp(\text{Energy}_{ijk'})}.
\]

The output is then computed as:
\[
\text{Output}_{ij} = \sum_k \text{Attention}_{ijk} \cdot V_{ik}.
\]

A residual connection and channel-wise concatenation are applied:
\[
\text{Enhanced\_features} = \gamma \cdot \text{Output} + x,
\]
where $\gamma$ is a learnable parameter. The enhanced features are merged using a $1 \times 1 \times 1$ convolution:
\[
\text{Final\_features} = \text{Conv3D}([\text{Enhanced\_features}; x]).
\]

This mechanism ensures that the model focuses on spatially significant regions, enabling precise segmentation of gliomas. The graph-based approach captures inter-voxel relationships effectively, improving segmentation accuracy while preserving the spatial and contextual integrity of the features.

\begin{algorithm}[h]
\caption{Graph Cross Attention}
\label{alg:graph_attention}
\begin{algorithmic}[1]
\Require Input tensor $X \in \mathbb{R}^{B \times C \times D \times H \times W}$
\Require Graph structure with edge indices $\text{edge\_index}$
\State Initialize Graph Attention Layers: $\text{GATv2Conv for } Q, K, V$
\State Initialize Softmax function $\text{Softmax}()$
\State Initialize learnable parameter $\gamma$
\State Initialize $1 \times 1 \times 1$ convolution $\text{Conv3D}()$
\State \textbf{Graph Construction:}
\For{\textbf{each} voxel $i$ in the 3D image \textbf{do}}
    \For{\textbf{each} neighbor $j$ of voxel $i$ \textbf{do}}
        \State Add edge $(i, j)$ to $\text{edge\_index}$
    \EndFor
\EndFor
\State \textbf{Forward Pass:}
\State Flatten spatial dimensions:
\[
Y \gets \text{Flatten}(X)
\]
\State Apply graph attention layers:
\begin{align}
Q &\gets \text{GATv2Conv}(Y, \text{edge\_index}), \\
K &\gets \text{GATv2Conv}(Y, \text{edge\_index}), \\
V &\gets \text{GATv2Conv}(Y, \text{edge\_index}).
\end{align}

\State Compute attention weights:
\[
\text{Energy} \gets Q \cdot K^T
\]
\[
\text{Attention} \gets \text{Softmax}(\text{Energy})
\]
\State Compute output features:
\[
\text{Output} \gets \text{Attention} \cdot V
\]
\State Reshape and merge features:
\[
\text{Enhanced\_features} \gets \gamma \cdot \text{Output} + X
\]
\[
\text{Final\_features} \gets \text{Conv3D}([\text{Enhanced\_features}; X])
\]
\State \Return $\text{Final\_features} \in \mathbb{R}^{B \times C \times D \times H \times W}$
\end{algorithmic}
\end{algorithm}

\subsection{Model Workflow}
The proposed solution integrates a modified U-Net architecture, MedNeXt, and a graph-based spatial attention mechanism to achieve accurate and interpretable segmentation of pediatric gliomas. The workflow comprises the following steps:
\begin{itemize}
    \item \textbf{Preprocessing:} 3D medical images are normalized, and robust data augmentation strategies are applied. Images are divided into smaller patches for efficient processing while preserving spatial context.
    \item \textbf{U-Net Encoder-Decoder:} A symmetric U-Net backbone with skip connections ensures spatial detail retention and reconstruction during segmentation.
    \item \textbf{MedNeXt Enhancements:} Incorporates ConvNeXt-inspired 3D convolutional layers tailored for volumetric data, improving feature extraction.
    \item \textbf{Graph-Based Attention Mechanism:} Implements the Graph Cross Attention (GCA) module at multiple decoder stages to enhance focus on diagnostically significant regions by capturing inter-voxel relationships.
    \item \textbf{Loss Function:} Combines cross-entropy loss for pixel-wise classification and Dice loss for overlap maximization, addressing class imbalance.
    \item \textbf{Post-Processing:} Smooths segmentation boundaries and reduces noise using morphological operations for better clinical usability.
    \item \textbf{Evaluation Metrics:} Utilizes Dice Score, IoU, and HD95 for quantitative performance evaluation.
    \item \textbf{Deployment Optimization:} Converts the trained model to ONNX or TensorRT formats, ensuring real-time inference capabilities in clinical environments.
    \item \textbf{Efficient Training:} Leverages GPU-accelerated training with FP16 precision, gradient accumulation, and advanced learning rate schedulers for handling large datasets effectively.
\end{itemize}

\subsection{Key Features}
\begin{itemize}
    \item {Modified U-Net Architecture:} Symmetric encoder-decoder structure with skip connections for effective feature retention and segmentation.
    \item {MedNeXt Architecture:} ConvNeXt-based 3D architecture with ReLU activation and multi-scale feature capture.
    \item {Graph-Based Spatial Attention:} Highlights tumor boundaries and distinguishes tumors from surrounding tissues with enhanced precision.
    \item {Advanced Loss Function:} Combines cross-entropy and Dice loss to tackle class imbalance.
    \item {Data Preprocessing and Augmentation:} Standardizes input data and applies augmentation for better generalization.
    \item {Batch-Based Processing:} Divides large scans into smaller batches to manage memory efficiently.
    \item {Post-Processing:} Reduces noise and smooths boundaries for clinical usability.
    \item {Evaluation Metrics:} Dice Score and HD95 ensure robust performance evaluation.
    \item {Clinical Deployment Optimization:} Converts the model for real-time inference using ONNX or TensorRT.
    \item {Efficient Training:} Utilizes GPU support, FP16 precision, and logging for handling large datasets efficiently.
\end{itemize}

\section{Results}

\subsection{Performance Metrics}
\subsubsection{Performance Metrics}
The Dice Score, also known as the Dice Similarity Coefficient (DSC), is a statistical measure widely used to evaluate the overlap between two sets, particularly in medical image segmentation. It measures how well the predicted segmented regions (e.g., tumor boundaries) align with the ground truth, which is often annotated by medical experts. Mathematically, the Dice Score is expressed as:

\[
\text{Dice Score} = \frac{2 \cdot |A \cap B|}{|A| + |B|}
\]

Here:
- \( A \): The set of pixels in the predicted segmentation.
- \( B \): The set of pixels in the ground truth segmentation.
- \( |A \cap B| \): The number of correctly predicted pixels.

The Dice Score ranges from 0 to 1, where:
- \( 1 \): Perfect segmentation, with complete overlap.
- \( 0 \): No overlap between prediction and ground truth.

The Dice Score directly quantifies segmentation quality by measuring both precision and recall:
\[
\text{Dice Score} = \frac{2 \cdot \text{Precision} \cdot \text{Recall}}{\text{Precision} + \text{Recall}}
\]
This property makes it particularly useful for imbalanced datasets, such as glioma segmentation, where tumor regions often constitute a small portion of the total image. Glioma segmentation requires accurate delineation of complex and irregular tumor shapes in 3D MRI scans. The Dice Score evaluates the agreement between predicted and ground truth regions, ensuring that segmented tumors align closely with their actual boundaries. High Dice Scores indicate accurate tumor detection and segmentation, which are critical for diagnosis, treatment planning, and monitoring disease progression. Conversely, low scores highlight model shortcomings and guide necessary adjustments.

\subsubsection{Hausdorff Distance (HD95): A Metric for Boundary Accuracy}
The Hausdorff Distance (HD) quantifies the maximum distance between the predicted and ground truth boundaries, providing a rigorous assessment of segmentation accuracy. The HD95, a robust variant, measures the 95th percentile of these distances, mitigating sensitivity to outliers while maintaining boundary precision.

Let \( S_P \) and \( S_G \) represent the sets of boundary points in the predicted segmentation and the ground truth segmentation, respectively. The directed Hausdorff Distance is defined as:
\[
H(S_P, S_G) = \max_{p \in S_P} \min_{g \in S_G} \| p - g \|,
\]
where \( \| p - g \| \) denotes the Euclidean distance between point \( p \) on the predicted boundary and point \( g \) on the ground truth boundary. The Hausdorff Distance is then:
\[
H_d(S_P, S_G) = \max \{ H(S_P, S_G), H(S_G, S_P) \}.
\]

The HD95 is computed as:
\[
\text{HD95}(S_P, S_G) = \text{Percentile}_{95} \{ \| p - g \| \mid p \in S_P, g \in S_G \}.
\]

Boundary accuracy is paramount in glioma segmentation, where tumor shapes are highly irregular and MRI scans exhibit significant intensity variations. The HD95 emphasizes the average worst-case boundary discrepancy, providing a robust measure of boundary alignment while ignoring extreme outliers.

This metric complements the Dice Score by offering a boundary-centric perspective. While Dice measures overlap, HD95 ensures that the segmented boundaries are spatially accurate, which is critical for estimating tumor size, shape, and proximity to vital structures in clinical applications.

\subsubsection{Computational performance}

In addition to segmentation quality, we evaluated the computational performance of the proposed model. Assessing CPU memory usage, GPU memory usage, and GPU power utilization provides insights into the model’s resource efficiency and its feasibility for clinical or large-scale research deployment.

\begin{itemize}
    \item \textbf{CPU Memory Usage:}  
    We monitored the CPU memory footprint throughout the training and inference processes. The proposed architecture maintained a steady and manageable CPU memory usage, ensuring that the model can be integrated into standard clinical workflows without requiring excessive computational infrastructure.

    \item \textbf{GPU Memory Usage:}  
    GPU memory utilization was recorded during training and inference. The model’s memory management strategies—such as batch sizing and model optimization techniques—enabled efficient use of GPU memory. This allowed for the processing of 3D medical images at a practical scale without resorting to high-end GPU systems.

    \item \textbf{GPU Power Utilization:}  
    GPU power consumption was tracked as a measure of energy efficiency. The proposed approach effectively balanced computational demands with resource use, minimizing unnecessary GPU load. An energy-efficient model not only reduces operational costs but also aligns with sustainable and environmentally-conscious computing practices.
\end{itemize}

\subsection{Performance Summary}
\begin{figure*}[htp]
    \centering
    \begin{subfigure}[b]{0.49\textwidth}
        \centering
        \includegraphics[width=\textwidth]{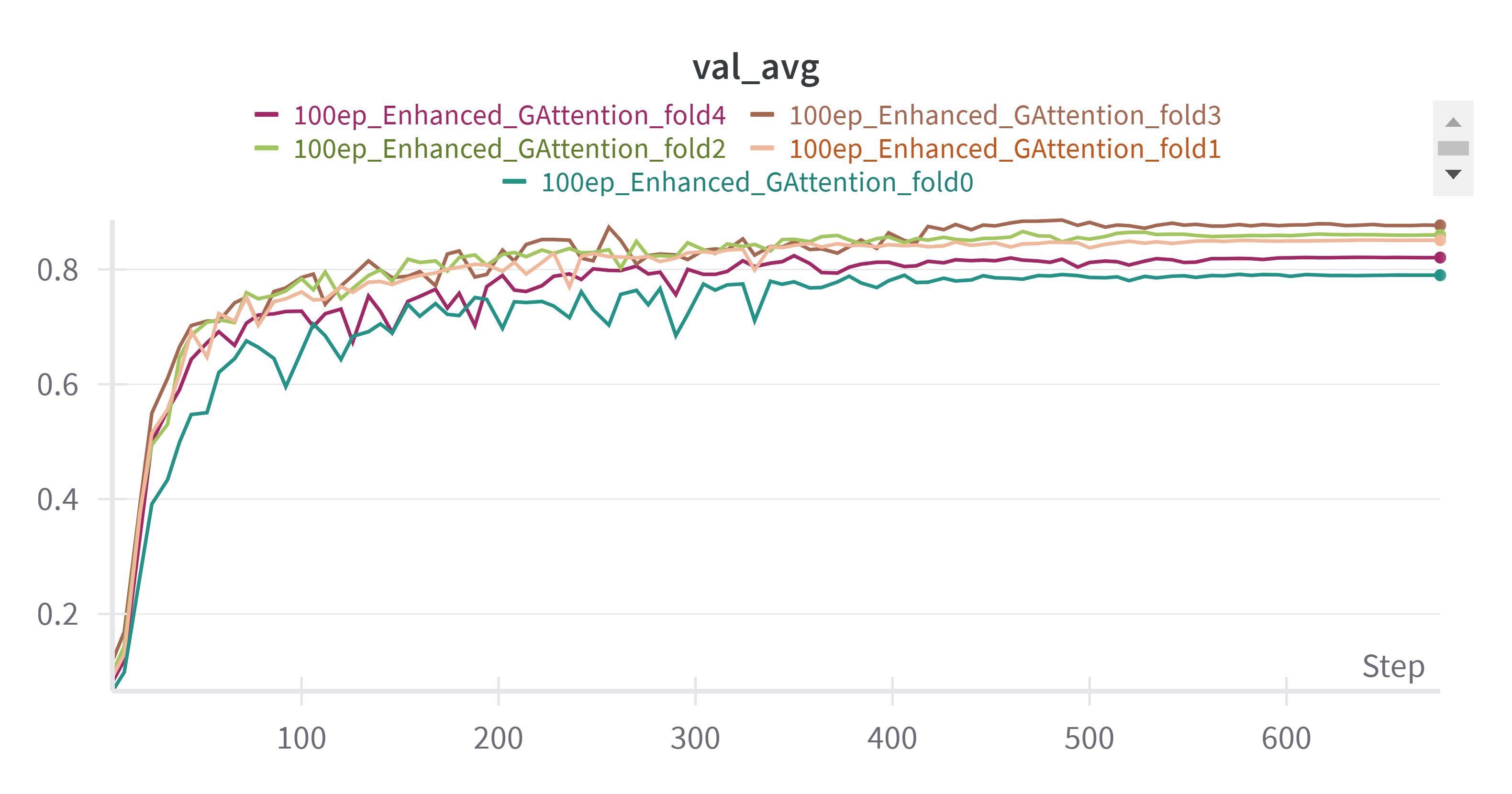}
        \caption{Average Validation Accuracy}
        \label{fig:val_avg}
    \end{subfigure}
    \hfill
    \begin{subfigure}[b]{0.49\textwidth}
        \centering
        \includegraphics[width=\textwidth]{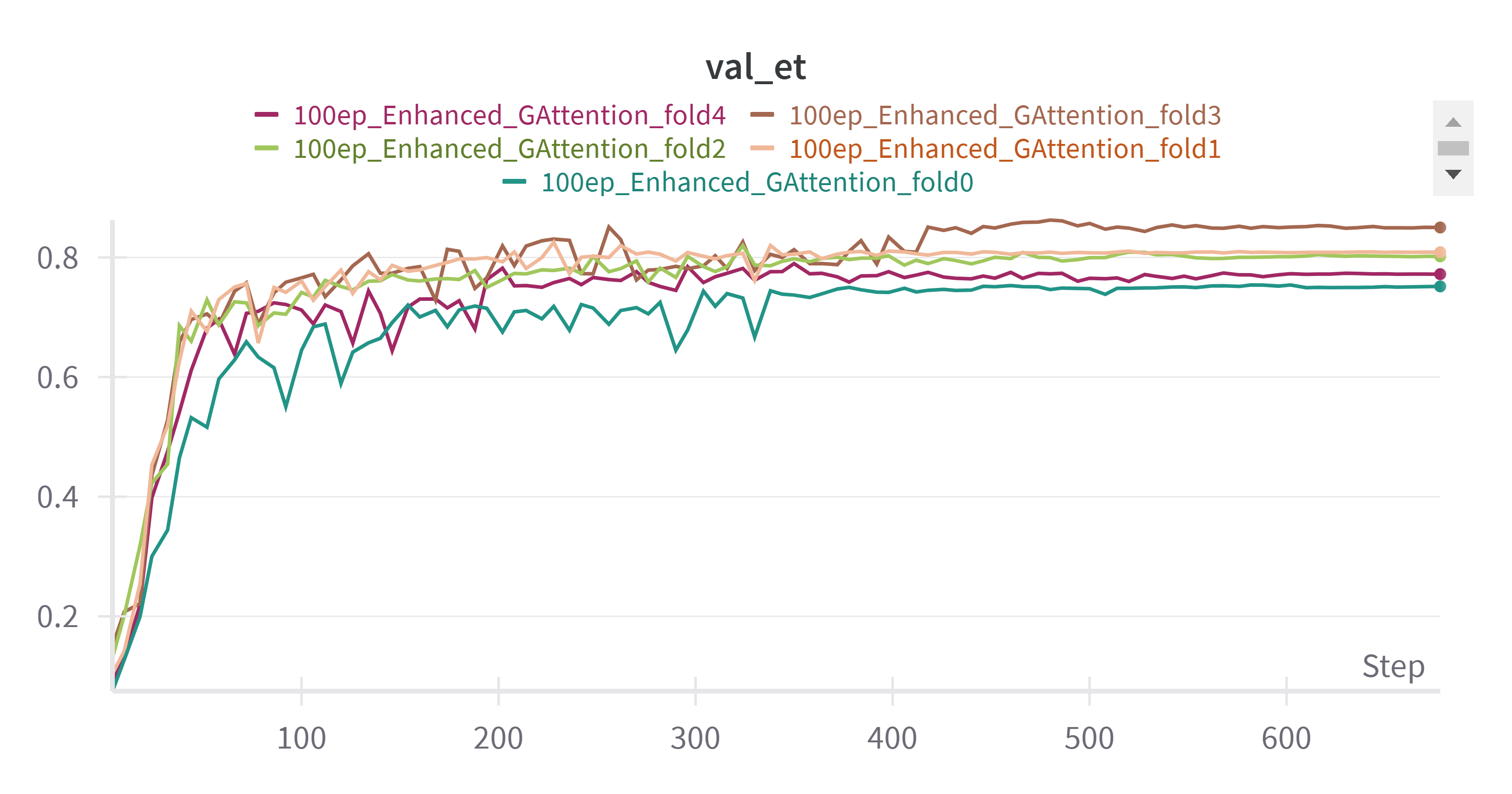}
        \caption{Enhancing Tumor (ET) Validation Accuracy}
        \label{fig:val_et}
    \end{subfigure}
    \vskip\baselineskip
    \begin{subfigure}[b]{0.49\textwidth}
        \centering
        \includegraphics[width=\textwidth]{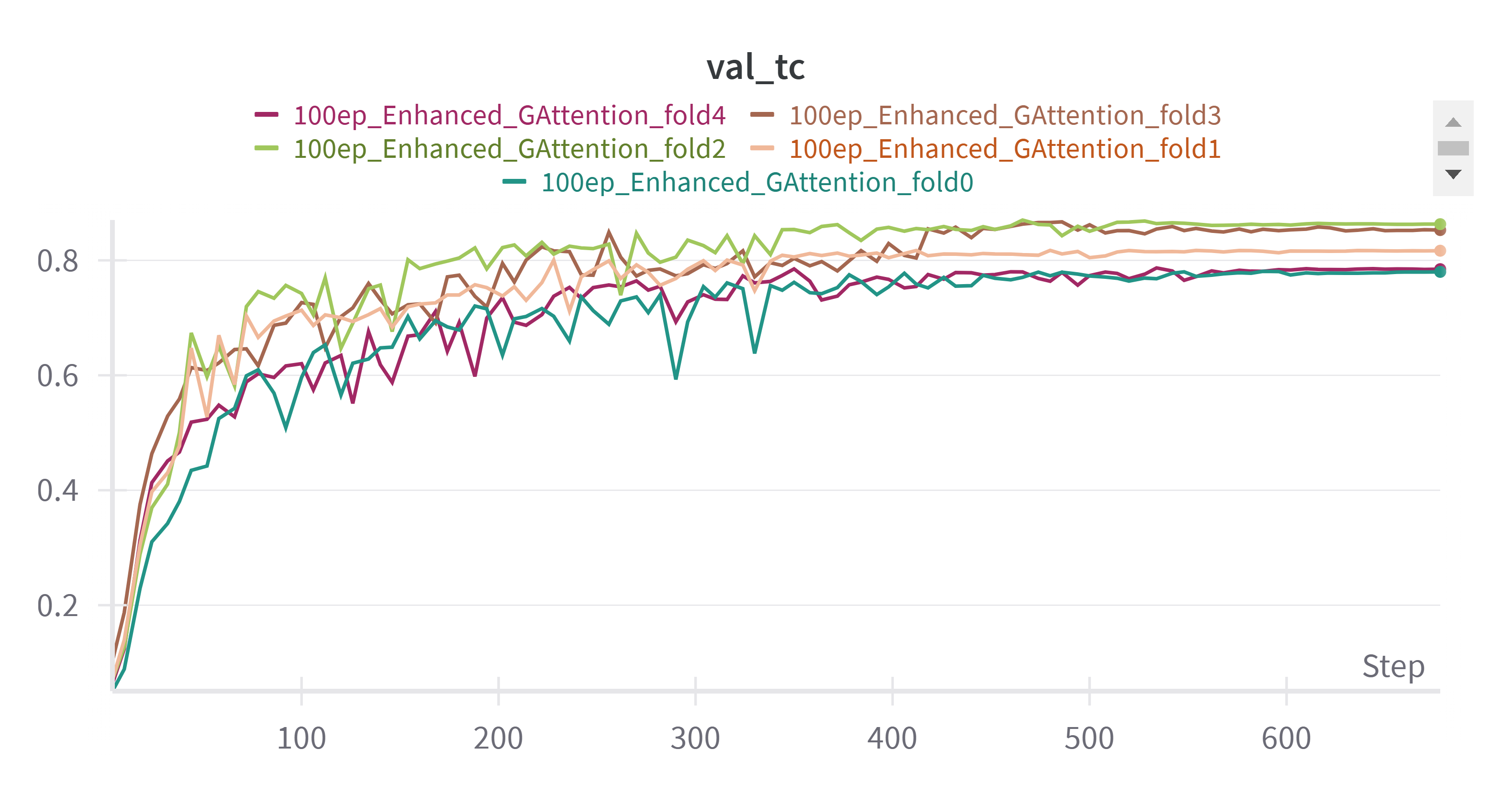}
        \caption{Tumor Core (TC) Validation Accuracy}
        \label{fig:val_tc}
    \end{subfigure}
    \hfill
    \begin{subfigure}[b]{0.49\textwidth}
        \centering
        \includegraphics[width=\textwidth]{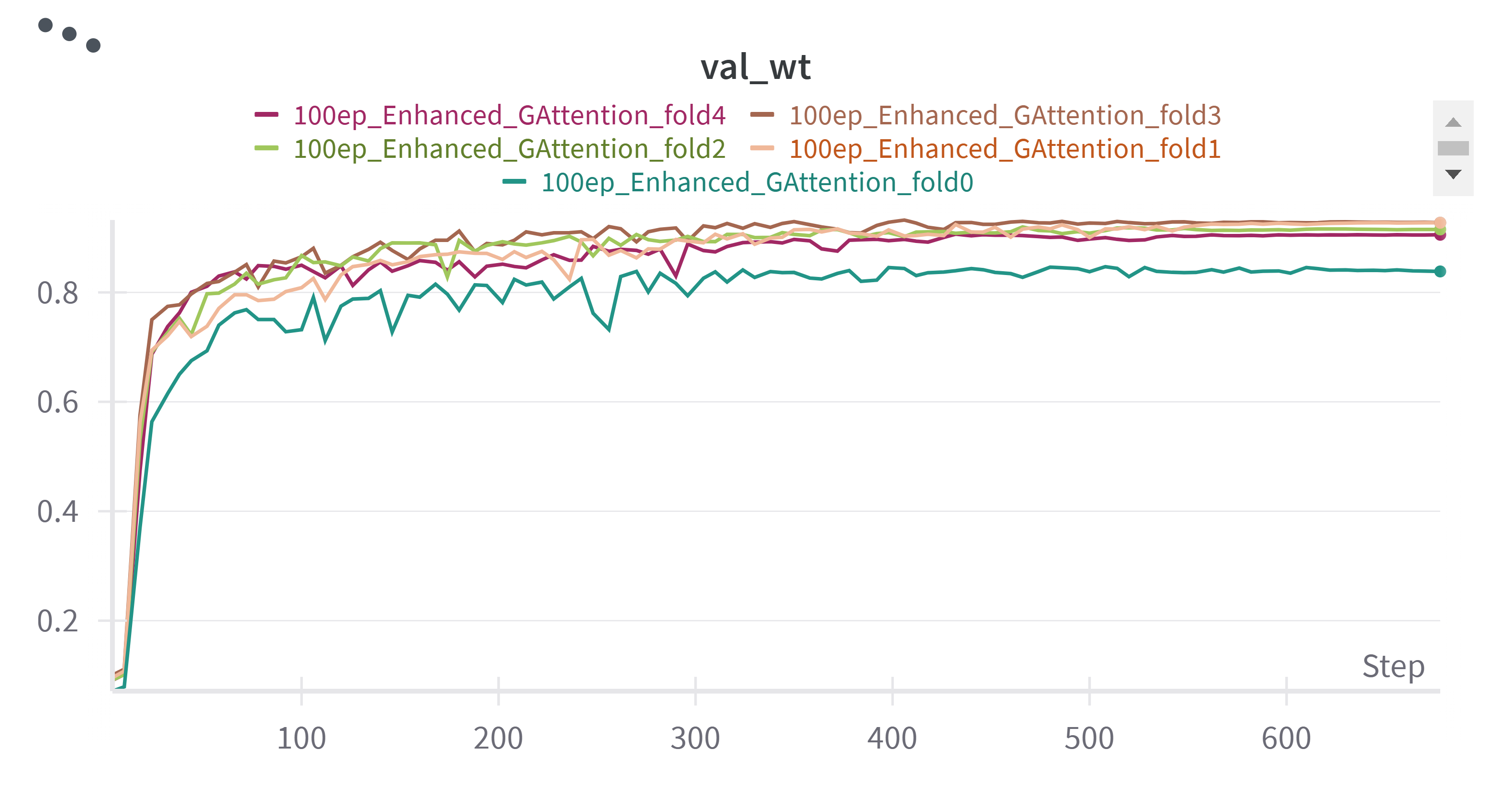}
        \caption{Whole Tumor (WT) Validation Accuracy}
        \label{fig:val_wt}
    \end{subfigure}
    \caption{Five-fold cross-validation accuracies for the proposed architecture, averaged and for each tumor sub-region.}
    \label{fig:acc_plots}
\end{figure*}

\begin{figure*}[htp]
    \centering
    \begin{subfigure}[b]{0.49\textwidth}
        \centering
        \includegraphics[width=\textwidth]{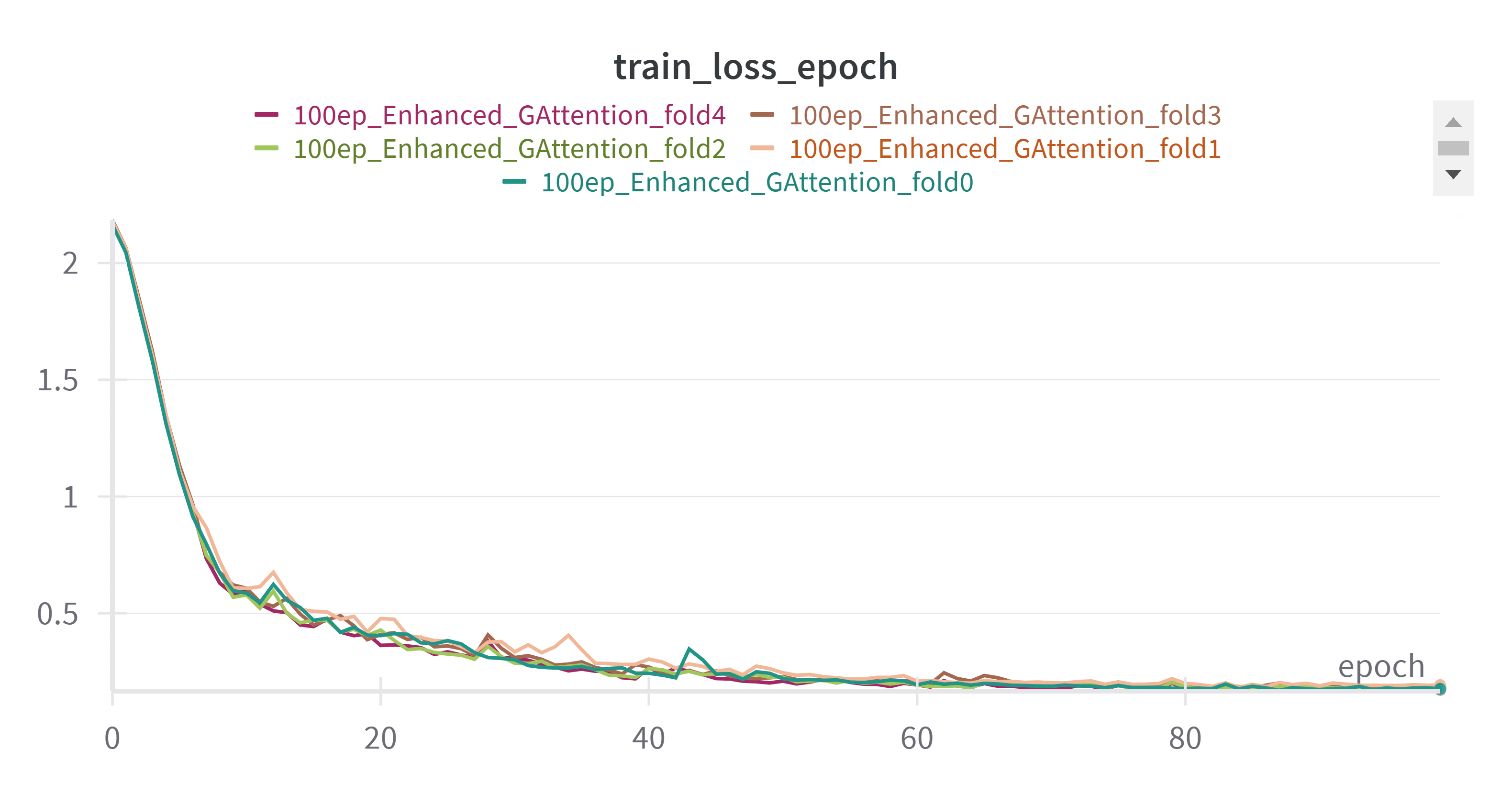}
        \caption{Training loss over 100 epochs}
        \label{fig:train_loss}
    \end{subfigure}
    \hfill
    \begin{subfigure}[b]{0.49\textwidth}
        \centering
        \includegraphics[width=\textwidth]{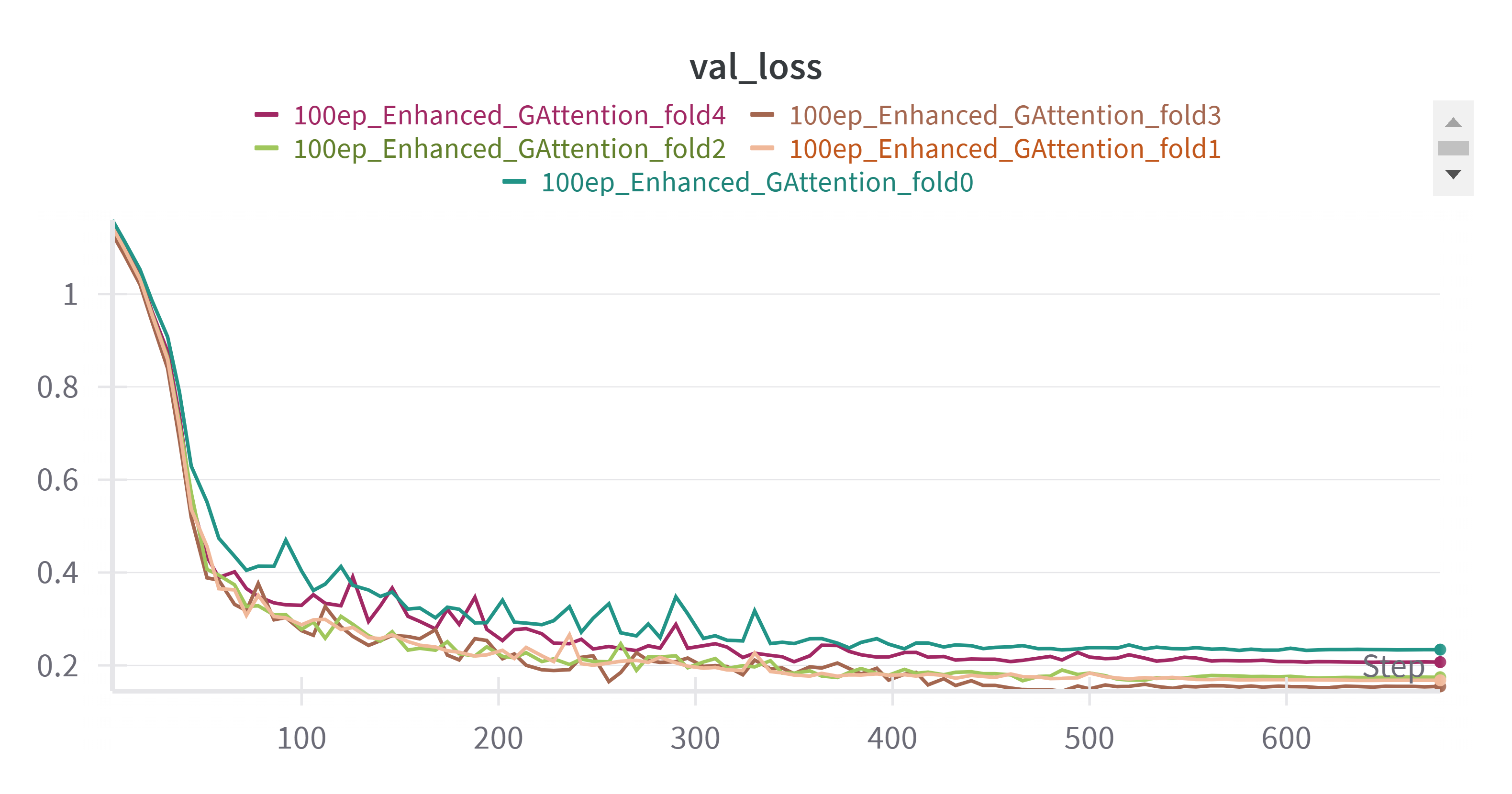}
        \caption{Validation loss over 100 epochs}
        \label{fig:val_loss}
    \end{subfigure}
    \caption{Five-fold cross-validation training and validation loss trends for the proposed model.}
    \label{fig:loss_plots}
\end{figure*}

Figures~\ref{fig:acc_plots} and~\ref{fig:loss_plots} illustrate the learning behavior of our proposed architecture across 5-fold cross-validation. Specifically, Figure~\ref{fig:acc_plots} shows both the average validation accuracy and class-specific accuracies for the Enhancing Tumor (ET), Tumor Core (TC), and Whole Tumor (WT) regions, while Figure~\ref{fig:loss_plots} presents the training and validation loss progression over 100 epochs. From these accuracy curves, we observe consistent improvements as training progresses, reflecting stable convergence. The model demonstrates strong performance across all tumor sub-regions, with the Whole Tumor accuracy typically converging rapidly and the Enhancing Tumor showing a more gradual improvement trend. Figure~\ref{fig:loss_plots} shows a steady decrease in both training and validation loss, indicating strong generalization capabilities. The close alignment between these curves suggests effective regularization and a minimal risk of overfitting.

\subsubsection{Quantitative Results}

\begin{itemize}
    \item \textbf{Dice Score:} \textbf{79.41\%}  
    This performance level indicates substantial overlap between the predicted and actual tumor regions. While it reflects strong volumetric agreement, it also suggests potential avenues for refining segmentation details to reach even higher accuracy.

    \item \textbf{HD95:} \textbf{12 mm}  
    With an HD95 of 12 mm, the predicted boundaries are generally within 12 mm of the ground truth. Enhancing this boundary precision could further improve the clinical utility of the segmentation, enabling more accurate treatment planning and monitoring.
\end{itemize}

\section{Conclusion}  
This study introduced an innovative approach to brain tumor segmentation, leveraging the MedNext architecture augmented with a 3D graph attention mechanism. The proposed model achieved a Dice Score of \textbf{79.41}, showcasing its ability to accurately delineate tumor regions within MRI scans. The integration of graph attention layers enabled the model to effectively capture spatial dependencies and intricate relationships across regions of interest, surpassing the performance of traditional convolution-based methods. These results demonstrate the potential of the proposed method to improve segmentation accuracy, offering a robust foundation for advancing automated tumor analysis.

\section{Future Work}  

Future work will focus on further optimizing the graph attention mechanism to enhance 
boundary precision and reduce segmentation errors, particularly around smaller tumor 
structures. Additional training on larger, more diverse datasets is planned to improve the 
model’s generalization capabilities across different patient demographics and scanner types. 
To facilitate clinical adoption, the model will also be integrated into a real-time processing 
pipeline, followed by a comprehensive evaluation in clinical settings. This approach aims to 
reduce manual segmentation workloads for radiologists and enhance consistency and accuracy 
in brain tumor diagnosis and treatment planning.

\section*{Acknowledgment}

We thank our mentor Dr. Sangeetha N whose guidance
 and feedback have been instrumental in shaping this research work and manuscript.
 This research was supported by our institution Vellore Institute of Technology,
 Chennai. We thank our academic department and colleagues for their insight
 and expertise that greatly assisted the research

\newpage
\bibliographystyle{ieeetr}
\bibliography{export}

\end{document}